\documentclass{./styles/svproc}
\usepackage{url}

\usepackage[acronym]{glossaries}
\newacronym{das}{DAS}{Driver Assistance Systems}
\newacronym{rl}{RL}{Reinforcement Learning}
\newacronym{vdc}{VDC}{Vehicle Dynamic Controls}
\newacronym{nn}{NN}{Neural Network}
\newacronym{drl}{DRL}{Deep Reinforcement Learning}
\newacronym{nmpc}{NMPC}{Nonlinear Model Predictive Controller}
\newacronym{nlp}{NLP}{Nonlinear Optimization Problem}
\newacronym{mpc}{MPC}{Model Predictive Controller}
\newacronym{cnn}{CNN}{Convolutional Neural Network}
\newacronym{cte}{CTE}{Cross Track Error}
\newacronym{ddpg}{DDPG}{Deep Deterministic Policy Gradient}
\newacronym{swa}{SWA}{Steering Wheel Angle}
\newacronym{tbp}{TBP}{Throttle/Brake Position}
\newacronym{a4ws}{A4WS}{Active Four Wheel Steering}
\newacronym{a4wd}{A4WD}{Active Four Wheel Drive}
\newacronym{as}{AS}{Active Suspension}
\newacronym{ppo}{PPO}{Proximal Policy Optimisation}
\newacronym{a3c}{A3C}{Asynchronous Advantageous Actor Critic}
\newacronym{ai}{AI}{Artificial Intelligence}
\newacronym{ml}{ML}{Machine Learning}
\newacronym{abs}{ABS}{Anti Block System}

\usepackage{cite}

\usepackage{graphicx}

\usepackage{array}

\usepackage{makecell}

\usepackage{ragged2e}

\usepackage[utf8]{inputenc}

\begin{document}
\mainmatter              
\title{Vision based driving agent for race car simulation environments}
\titlerunning{Race Car Self Driving}  
%
\author{Gergely Bári \and László Palkovics}
\authorrunning{Gergely Bári et al.} 
%
%
\institute{Széchenyi István University, H-9026 Győr, Hungary\\
\email{gergely.bari@humda.hu}} 

\maketitle              

\begin{abstract}
In recent years, autonomous driving has become a popular field of study. As control at tire grip limit is essential during emergency situations, algorithms developed for racecars are useful for road cars too. This paper examines the use of Deep Reinforcement Learning (DRL) to solve the problem of "grip limit driving" in a simulated environment. Proximal Policy Optimization (PPO) method is used to train an agent to control the steering wheel and pedals of the vehicle, using only visual inputs to achieve professional human lap times. The paper outlines the formulation of the task of time optimal driving on a race track as a deep reinforcement learning problem, and explains the chosen observations, actions, and reward functions. The results demonstrate human-like learning and driving behavior that utilize maximum tire grip potential.
\end{abstract}

\section{Introduction}
Race cars always corner on the tire grip (friction) limit, and vehicle control (driving) on the tire grip limit is important during accident avoidance for self-driving passenger cars. This is a good reason why the development of an algorithm that can drive a real vehicle at the level of a racecar driver is an interesting research topic. This paper aims to demonstrate that state-of-the-art \acrshort{rl} methods are indeed useful tools for solving this problem, as a fist step, in a simulated environment.

Most articles in this field address the development of classic subsystems according to the problem decomposition, such as trajectory planning or vehicle control. There are less learning-based approaches documented yet; however, their advantages seem to be well documented in the past. \cite{remonda_acting_2021,remonda_formula_2021,fuchs_super-human_2020,wurman_outracing_2022}

In \cite{fuchs_super-human_2020} the task of autonomous car racing is solved in a computer game called Gran Turismo Sport.  The results show that the obtained controllers not only beat the built-in nonplayer character of the game, but also outperform the fastest known times in a dataset of more than 50,000 human drivers. In the currently available literature, this is the closest approach to that taken in this work. Although the key ideas are similar, the use of visual input in this work introduces significant differences between the two works.

\section{Modelling approach}
\label{sec:methods}
\begin{figure}[!h]
 \centering
 \includegraphics[width=7cm]{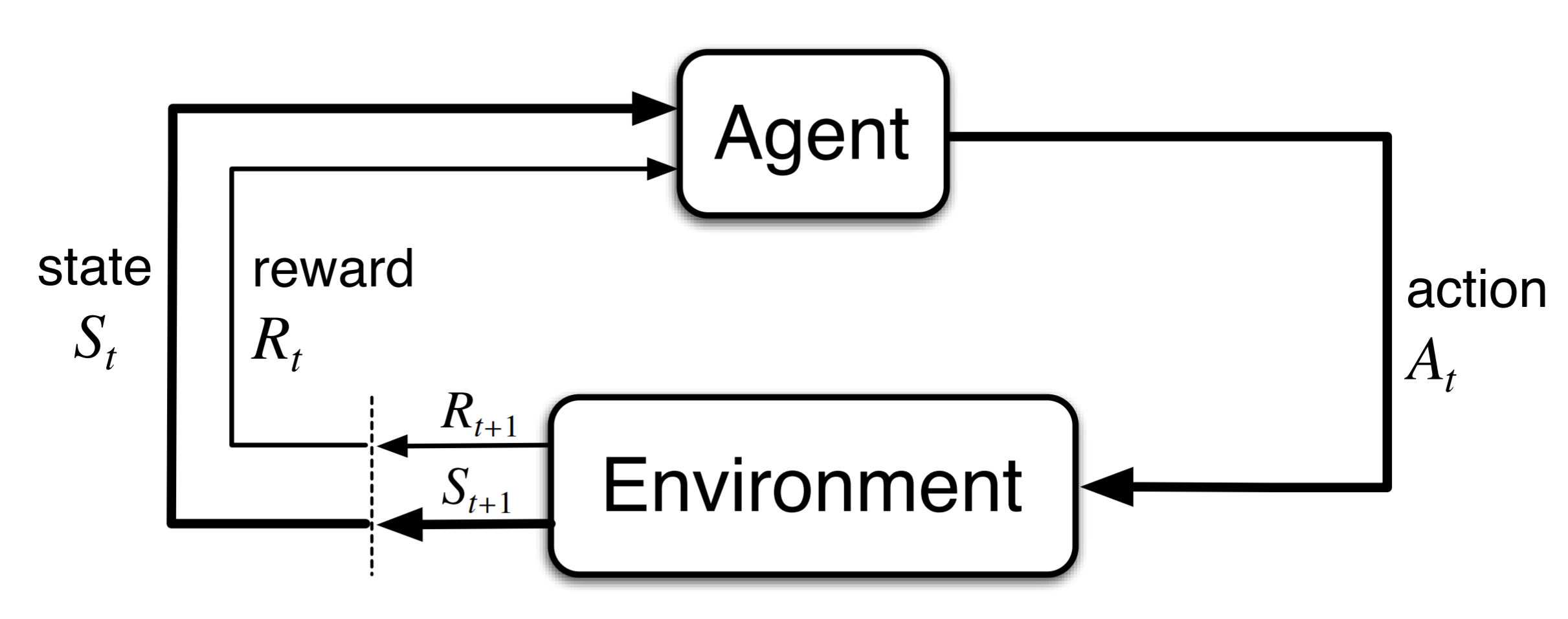}
 \caption{Scheme of the so called Markov Decision Process, formalizing the Agent - Environment interaction in Reinforcement Learning problems \cite{sutton_reinforcement_2018}}
 \label{fig:RL}
\end{figure}

Considering all above, the chosen goal in this work was to use state-of-the-art \acrshort{rl} methods, and create an agent, capable of driving in simulation a specific racecar, on a specific racetrack with lap times of a professional human driver, while using only visual information (pixels).
The task of driving a race car is transformed into a \acrfull{rl} problem following the general \acrshort{rl} approach pictured in Figure \ref{fig:RL}.

In this work, the \textbf{ agent} is a single \acrfull{nn}. This \acrshort{nn} has a special type, it is called \acrfull{cnn}. The \acrshort{cnn} used here has a structure presented in \cite{mnih_playing_2013}. 
The \textbf{Action} decided by this agent consists of two real numbers in the range of [-1..1]. One of these represents the scale of maximum throttle (+1) to maximum brake, while the other represents the steering wheel angle from maximum left (+1) to maximum right (-1). It is common in such works to limit the agent's control abilities to human level, which usually means the introduction of some kind of rate limit on the actions. In this work, the agent-to-environment interaction happened at 20Hz, while the human professional race car driver's control frequency is generally considered around 10Hz. As the main goal of this work was to prove that RL algorithms can indeed create professional level race driving behavior, and not a clear human vs. AI competition, this difference was accepted.
The used \textbf{Environment} is an open-source race car simulation software called TORCS \cite{noauthor_torcs_2020}. TORCS, being a detailed car simulation, can provide vehicle dynamic signals and visual 3D representation (pixels) about the vehicle-track environment. In this work these pixels are used as the \textbf{State}, while the vehicle dynamic signals are used to form appropriate Reward for training the \acrshort{rl} algorithm.
This work uses a single \acrshort{nn} as agent, so we use a so-called end-to-end method, without further decomposition of the sensor-to-actuator mapping. Although the choice of usual Perception-Planning-Control subsystems seems advantageous in case of road cars, it can lead to some problems in case of racing.
In case of road cars, safe trajectories can be planned assuming some worst-case grip values, while in racing, planning a time optimal trajectory would require estimating the grip with very high ($<1\%$) precision, and currently available grip estimation techniques are only in the range of $\sim10\%$ accuracy \cite{acosta_tire_2018}.

To put the grip sensitivity in perspective, the $1\%$ change in grip level can easily result in $~0,15\%$ lap time difference \cite{kelly_time-optimal_2010}. In numbers, this means that on a 1minute 40second lap, dropping average grip level from 1.0 to 0.99 will result in 0.15sec increase in lap time, which can easily decide between first and second places.
Even if the grip levels (tire model parameters) are completely known, the computational needs to establish the ideal racing line are usually high, or the results are not comparable with those of a human driver \cite{perantoni_optimal_2014}. As the grip level cannot be estimated with a good enough precision, the tire model, and hence the whole vehicle model cannot be known too. Therefore, it is reasonable to try the so-called model-free techniques for driver modeling in such setting.
This kind of uncertainty and sensitivity transforms the race driving behavior modeling from control design task into a risk management task. The racing driver does not have a clear trajectory in mind during driving. It is continuously trying to find the proper direction and maximum magnitude of the in-plane acceleration of the vehicle, and the trajectory evolves while the boundary of the vehicles' acceleration capabilities is tracked. During this boundary tracking, there is a certain level of risk arising from the fact that the exact grip (the potential of the vehicle to change state) is unknown and the car can slide off the track when crossing its limits.
This kind of risk management appears in the evolution of lap times during a qualifying session, too. In these cases, drivers usually have 3 – 4 attempts to make their quickest lap times, and it is usual that we see laps getting faster and faster from attempt to attempt. There can be technical reasons for this (e.g. track grip evolution, vehicle mass reduction with fuel use, etc.) but the fact that drivers accept higher and higher chances of making mistakes is another key reason for this. Managing this risk is also important to have better and better lap times, and it is done by driver intuition during race car driving.
As reports about recent \acrlong{drl} achievements usually report some kind of emergent intuition of these agents \cite{silver_mastering_2016}, so it is straightforward to try modeling this risk management with such deep learning techniques instead of traditional ones.

In \acrlong{rl} \textbf{Reward} is a special signal with a numerical value that the agent seeks to maximize over time through its choice of actions. In the present work, the goal is to create an agent with the behavior of minimizing lap time. An important aspect of constructing a reward function is how dense or sparse it is. Using only lap-time as a reward signal is a very sparse reward. It makes learning very unstable and slow, as the feedback about actions comes only after one lap (end of episodes), which can easily be a few thousand steps.
To construct a reward that is available at each step (dense), we start by considering how race car drivers learn to drive fast. For race car drivers, the main feedback signal when practicing is a so-called time-difference signal. Usually this value is calculated for every moment and displayed on the driver's dashboard. This signal compares the actual lap to a previous "reference lap" (usually the fastest lap of the driver) and shows the driver how much shorter time it took to reach the actual track position in the actual lap compared to reach the same position in the reference lap. So when drivers experience with a different strategies, (eg.: different racing lines in a corner), they tend to check if the time difference on their dashboard increased or decreased during the corner. Based on these, chosing the change in the time-difference signal as the basis of the reward function seems a good choice for the purpose of this work.
Considering a reference lap that is basically a constant speed movement along the track centerline, and assuming fixed time steps between each Agent-Environment interactions, this reward simplifies to the progress made in each time step. It worth noting that this thinking, in principle, leads to the same reward used in \cite{fuchs_super-human_2020,wurman_outracing_2022}. 

\begin{equation} \label{eq1}
r^{t_{diff}}_{t} = s^{cl}_{t} - s^{cl}_{t-1}
\end{equation}

where $s^{cl}_{t}$ is the distance traveled along the track centerline in timestep $t$.

To make training more robust, additional components were introduced into the final reward value. One component was defined depending on the reason why a learning episode was terminated ($r^{ter}_{t}$). In this, a constant value was added to the reward if the agent finished the lap and penalties were introduced for the reasons: leaving the track, "turning back" on the track, if the car damaged, progressed "backwards" on the track, or the agent made too small progress in a given time window. In addition, a reward component was added to punish the agent using "too high" action values ($r^{act}_{t}$). As the neural network output can be any value, it was found during training that adding a reward component that punishes non-feasible, out-of-bound actions helped stability of training.

\begin{equation}
  \label{eq:actbound}
  r^{act}_{t}= (\frac{|a_{t}|}{p^{sc}} - p^{bnd} + 1)^{2}
\end{equation}

The final reward function is then:

\begin{equation} \label{eq2}
r_t = r^{t_{diff}}_{t} + r^{ter}_{t} - r^{act}_{t}
\end{equation}

Specific values for these parameters are summarized in Table \ref{table1}.

For training the agent the \acrfull{ppo} algorithm was used. \cite{schulman_proximal_2017} \acrshort{ppo} is a well-known, widely used on-policy, model-free \acrshort{rl} algorithm. In recent years, it has proved its robustness in various use cases. \cite{openai_dota_2019, openai_solving_2019}
For gathering experiences for training multiple ($32$) TORCS instances were run in parallel. Trainings with various hyper parameters were performed on DELL R730 E5-2670 v3, computers. Trainings were stopped after a maximum of 10 days, which corresponds approximately one billion training steps in this case. The performance of the agent was monitored during the training and in every $10 000$ step a test episode was performed. In this case the mean value of the agent \acrshort{cnn} stochastic output (actions) was used, while during learning actions were sampled from these stochastic variables. Training parameters for the results presented in this work are summarized in Table \ref{table2}.

\section{Simulation Results}
\begin{figure}[!h]
  \centering
  \includegraphics[trim = 23mm 17mm 15mm 25mm, clip, 
   width=11cm]{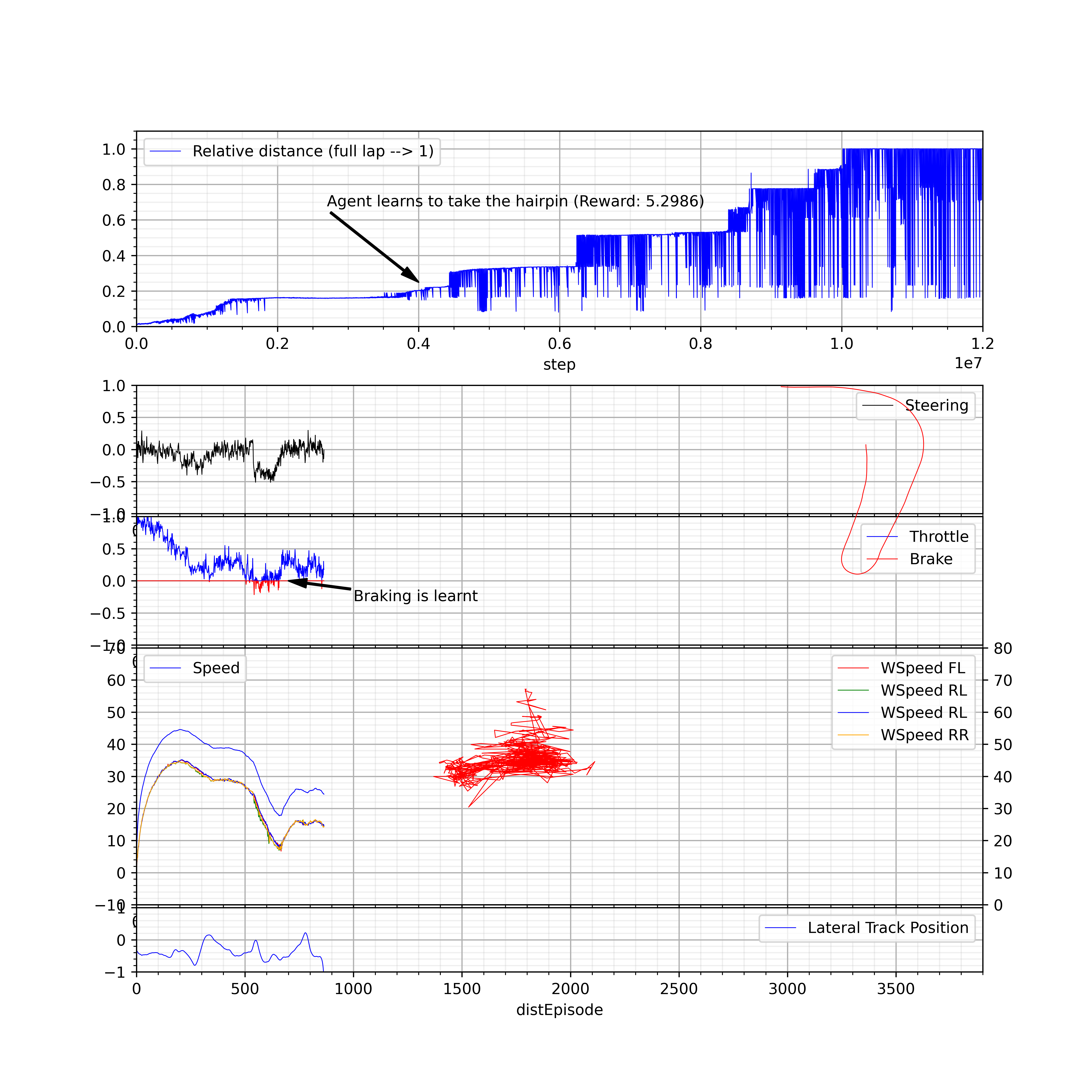} 
  \caption{An episode showing that the Agent learns breaking and takes the hairpin, but fails to turn left in the following (first left) corner}
  \label{fig:HairpinOKf1a}
 \end{figure}

\begin{figure}[!h]
 \centering
 \includegraphics[trim = 23mm 17mm 15mm 25mm, clip, 
  width=11cm]{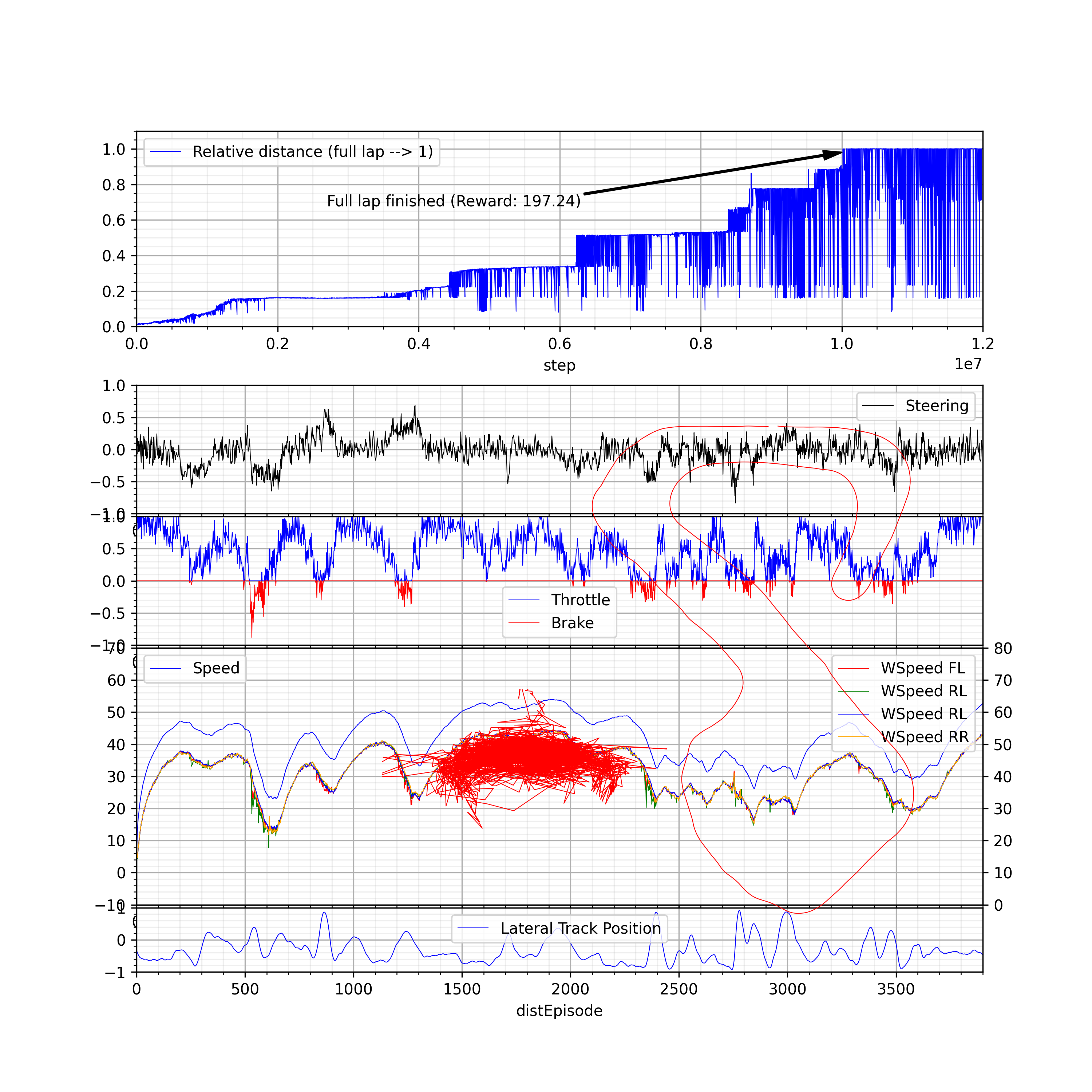} 
 \caption{An episode, showing Agent learned to drive the full lap for the fist time}
 \label{fig:fulllapf1a}
\end{figure}

\begin{figure}[!h]
 \centering
 \includegraphics[trim = 23mm 17mm 15mm 25mm, clip, 
  width=11cm]{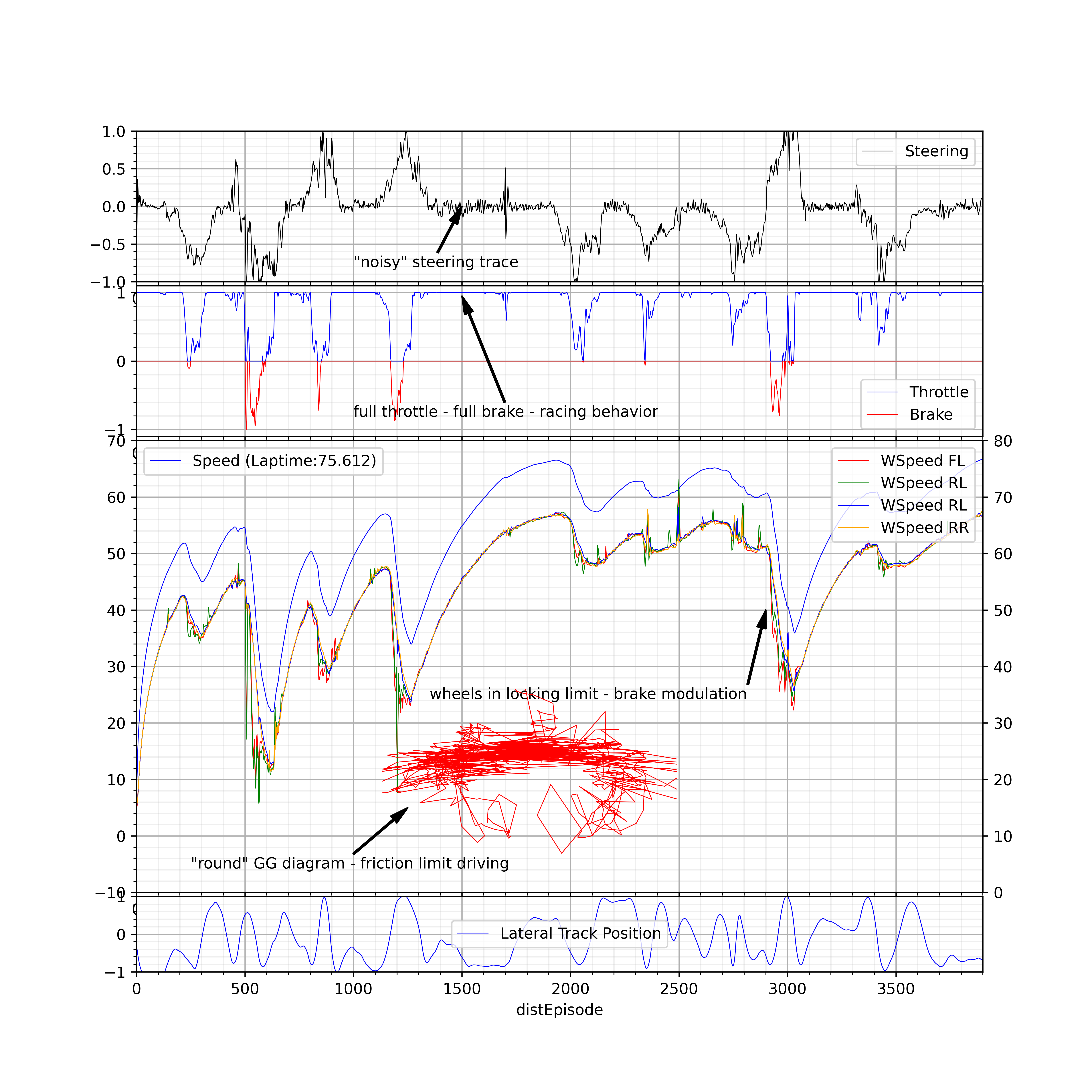} 
 \caption{A plot showing human pro-like driving behaviour.}
 \label{fig:fastlapf1a}
\end{figure}

To efficiently show important aspects of the results, a specific figure layout is used in this section, so Figures ~\ref{fig:HairpinOKf1a}, ~\ref{fig:fulllapf1a}, and ~\ref{fig:fastlapf1a} share the same structure. The top graph shows how far the agent can drive on the track as a function of the learning steps, while the bottom graphs show signals as a function of the traveled distance. This second part shows on the top the "driver inputs", eg.: the steering and the throttle/brake signals (+1 means full throttle, -1 means full brake pedal application), then the graph below shows vehicle speed with wheel speeds in [m/s] (y axis is in offset), and the last graph on the bottom shows track position, where +1 refers to the left, -1 the right side limit of the racetrack. There is also a red "patch" in the middle of the speed trace. This patch is created by plotting the longitudinal and lateral acceleration of the vehicle. In racing terms, this is the so-called "GG diagram". Having a nice "round" or "fat" GG diagram generally shows that the car is driven close to its limits, while a GG diagram with a "cross-like" or "thin" shape usually refers to a more slow, road-car-like driving behavior. To better understand the location of the track, the plots also show the traveled trajectory of the vehicle, as an overlayed single line over the steering, throttle / brake, and speed diagrams.

Several learning runs were performed in this work. Trials with various reward functions or hyperparameters resulted in more than a hundred runs, and most of these runs lasted hundreds of millions of learning iterations. However, in all of them, the agent learned relatively quickly in $\sim$10 million steps how to complete a full lap, and during this phase the learning showed the same pattern. The interesting aspect of this pattern is described in the following.


In Figure~\ref{fig:HairpinOKf1a} it can be seen that at the beginning of the training process the unlearned agent shows random behavior and shortly after starting the episode it leaves the track. This is represented in the top graph, as ~0 learning steps corresponds to ~0 distance. 
As learning steps grow, the distance traveled also increases. This shows how the agent learns to drive straight. The first "plateau" in the traveled distance starts at $\sim$ 0.5 million steps, slightly less than 20\% of the total lap distance (the relative distance is $\sim$ 0.2). Here, the agent learns to steer according to the first (almost straight, full throttle, relatively "easy") corner, while at $\sim$4.5 million steps and slightly more than or 20\% of the total lap distance, the braking behavior emerges. Here the agent reaches a hairpin, that is a small radius, long-arc corner, which can only be performed with small speed. Therefore, the agent must learn that at some point it needs to apply brake instead of throttle to keep the car on track. Note that this plateau is quite long. It took $\sim$4 million steps ($\sim$ 40\%) to learn breaking and take this single corner.

Figure~\ref{fig:fulllapf1a} shows an episode created at $\sim$10 million learning steps, where the agent learned to go around the full lap for the first time. Keeping focus on the top graph, the next "step change" in the traveled distance comes $\sim$6.2 million learning steps. Here, the agent learns to turn left and progresses to the next sequence of corners. Up until this point, the track had only right corners (first corner and the hairpin), so taking these two left-handers required learning a new behavior. Note that until $\sim$8.3 million steps, the agent does not learn to process further. Here, the agent has already spent 80\% of its "full lap learning time" ($\sim$8 of $\sim$10 million steps), but can only reach $\sim50\%$ of the lap yet. This pattern of learning makes perfect sense. After long learning to accelerate, brake, turn right and left, the agent will learn faster the remaining corners, almost a quarter of the steps ($\sim$2 million steps).
The next notable milestone is how the agent learned a fast right-hand sequence of the track.($\sim$8.5 million steps) These corners are very fast with lots of bumps. This is a very tricky part of the track, as the fast speed combined with the bumpiness makes the car unstable. So although the agent has already learned to turn right in the hairpin before, these right corners require totally different behavior. It is also noteworthy that the agent basically learned this sequence at once; after it learned to take the first of these kinds of turns, it passed all the others too with almost no learning.

The agent also needs to take some steps to learn the last left-hand turn after the quick right corner sequence. ($\sim$ 9.5 million steps) Although this corner is a left-hander and seems the same type as the previous two left-handers, this corner is still a "new" type thanks to the preceding quick right turns that affect the ideal line for this corner. The short straight before this corner requires to start braking approximately in the middle of the track instead of the usual choice of the outside edge. Consequently, even though this is not the first left corner to be learned, the approach is still novel and, as a result, it takes approximately one million steps to learn this turn, which is a considerable amount of time at this relatively late stage of the learning process.

In the final phase of this learning, at $\sim$10 million learning steps, the agent can drive around the lap, however, the plots in Figure ~\ref{fig:fulllapf1a} show that it is not yet driving at the grip limit. One clear sign of this is the pattern in the throttle/brake input curve. Racing-like driving usually means constant full throttle application in straights, with sudden change to high brake pedal application before corners that is not seen here yet.  Furthermore, the GG diagram (red patch) shows low braking dynamics because the shape is not round.
\begin{figure}[!h]
 \centering
 \includegraphics[trim = 23mm 50mm 23mm 12mm,clip, 
  width=11cm]{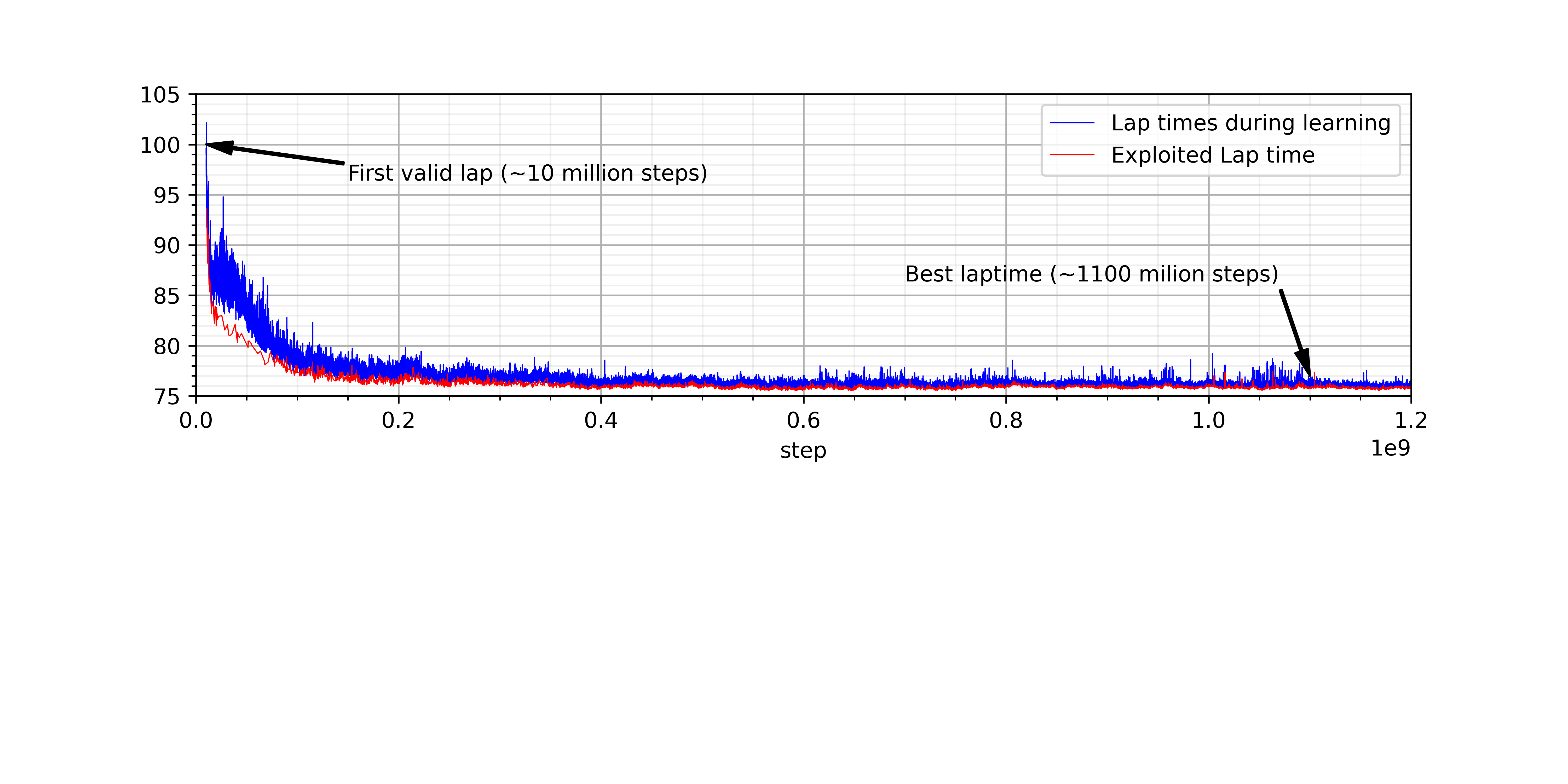} 
 \caption{Lap time evolution during learning}
 \label{fig:Learninglaptime}
\end{figure}

After learning to drive around the track in $\sim$10 million steps, it took 100 times more steps ($\sim$ 1000 million steps) for the agent to evolve to the professional human level. Figure~\ref{fig:Learninglaptime} shows this disproportion nicely.

Figure~\ref{fig:fastlapf1a} shows a driving behavior that is very similar to that of a professional race car driver. The patterns on the graphs are almost identical to those recorded from real race car drivers, which is further corroborated by interviews with experts (race engineers and professional drivers). The only trace that is noticeably different from that of human drivers is the steering wheel angle trace. This is to be expected, as the agent in this work was not subject to any rate limit, and the power needed to turn the steering wheel was not modeled. Considering throttle/brake signal, the pattern here is indistinguishable from a human race driver based on human experts feedback. In a straight line there is full throttle application (flat sequences at maximum (+1)) When the agent reaches the braking point before a given corner, this maximum throttle turns into maximum brake signal (maximum as maximum possible brake without locking the wheels). Note that this is also a common pattern in classical time-optimal control problems. The GG diagram in Figure~\ref{fig:fastlapf1a} is also a key indicator of grip limit maneuvering, as it has a nice round shape in this case.

Note one really interesting aspect of this behavior. The agent learns how to brake on the tire grip limit without blocking the tires, although only visual information is available, without direct wheel speed signals. This is clear by checking the wheel speed traces in Figure~\ref{fig:fastlapf1a}. There are noise-like spikes during braking on the wheel speed traces. This indicates that the wheels are in the locking limit, but none of them actually stops rotating (drops to 0 speed). This pattern in wheel speed then shows up in the brake pedal usage too. Where the wheels show a locking pattern (some wheels start to slow down more than others), the agent decreases the braking effort.

Learning this antilock-like behavior without direct wheel speed information is not straightforward, as wheel speed is generally considered essential for \acrfull{vdc} functionalities like \acrfull{abs}. 
One possible reason for this is that the agent has not direct (wheel speed) but indirect information about wheel locking. It is an essential property of tires, that when they are locked they lose the capacity to create lateral forces, thus control through steering is lost in such situation. Therefore, although the agent does not know the rpm of the wheels, it can recognize the loss of steering control when the wheels are blocked. As the expected episode reward will be less in these cases, this feeds back into learning to avoid these situations by modulating brake pressure to avoid wheel lockups.

\begin{figure}[!h]
 \centering
 \includegraphics[trim = 23mm 20mm 15mm 25mm, clip, 
  width=11cm]{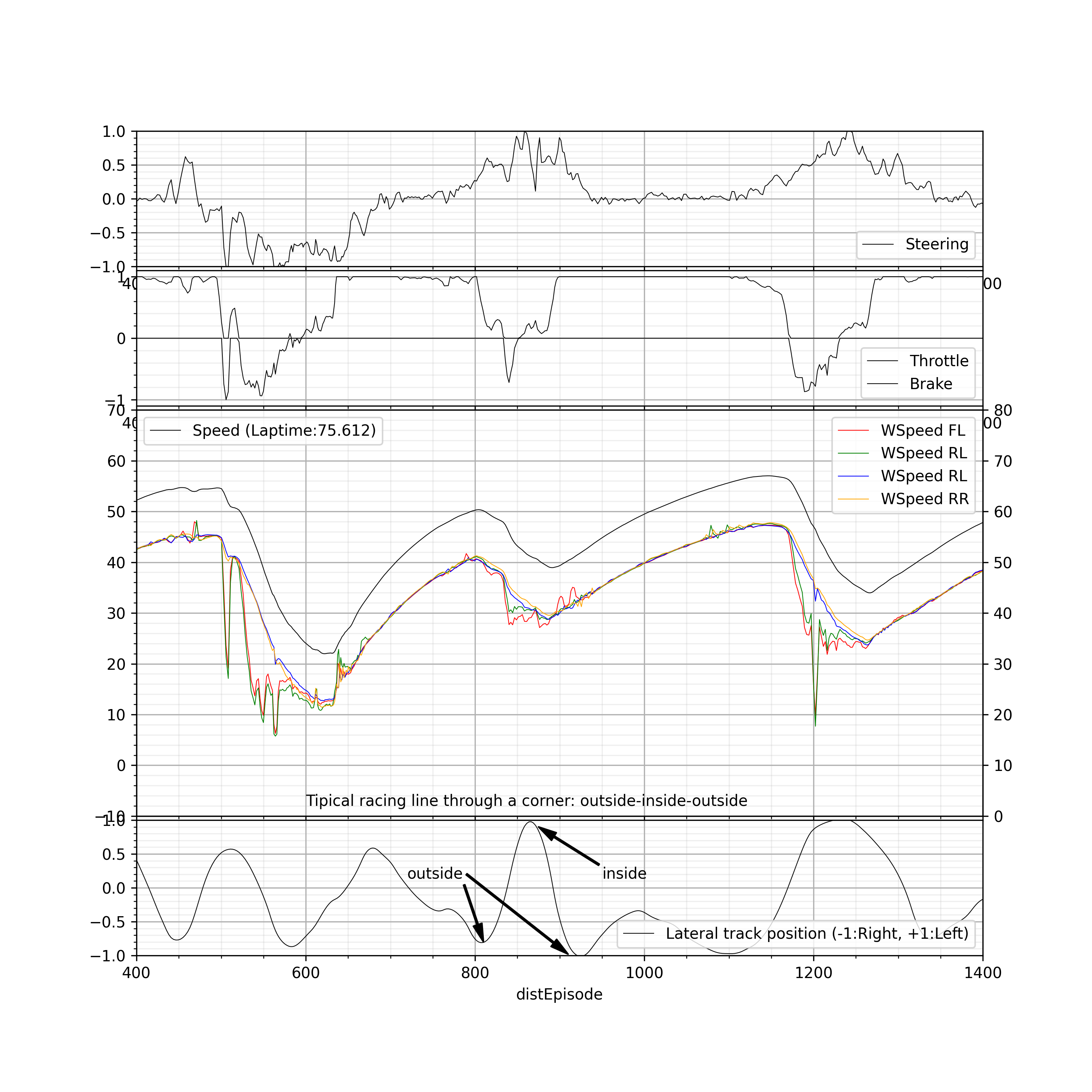} 
 \caption{Plots showing how the agent chooses racing line in a corner}
 \label{fig:outinoutf1a}
\end{figure}

An other interesting result is that the agent learns to select an optimal racing line through the corners, as seen in the bottom graph of Figure~\ref{fig:outinoutf1a}. (This is known as "trajectory planning" in classical self-driving terminology). It can be seen that before the agent starts braking, it approaches the outside edge of the track. Then it approaches the inside edge, approximately when it reaches the minimum speed (in racing terms: it reaches the apex), and lastly it "falls" to the outside edge of the track again when it applies maximum throttle. 

\section{Conclusion}
This work demonstrates the effectiveness of deep reinforcement learning algorithms in learning self-driving behaviors without prior knowledge or experience. Results show that self-driving on the tire grip limit can be formed as a reinforcement learning problem and that Proximal Policy Optimization is suitable to solve this problem using visual state information.

It would be interesting to investigate the effects of allowing an AI agent to control all four wheels independently, with the ability to adjust the torque and steering angle of each wheel, as well as to compare the agent's performance with a human driver in both simulated and real world scenarios. Additionally, it would be beneficial to research the behavior of the agent if the controlled car has Anti-lock or Traction control systems


\section{Acknowledgement}
The research was supported by the European Union within the framework of the National Laboratory for Autonomous Systems. (RRF-2.3.1-21-2022-00002)

%
%
\bibliographystyle{./styles/bibtex/splncs_srt}
\bibliography{GBari_VisionBased}

\section{Appendix}

\begin{table}[!h]
\caption{Parameters for the reward function in eq(\ref{eq2})}
\begin{center}
\begin{tabular}{r@{\quad}rl}
\hline
\multicolumn{1}{l}{\rule{0pt}{12pt}
                   Parameter}&\multicolumn{2}{l}{Value}\\[2pt]
\hline\rule{0pt}{12pt}
Reference Speed ($v^{ref}$) & $20$ \\
Action scaling (\texttt{$p^{sc}$}) & $15$\\
Bound for scaled action (\texttt{$p^{bnd}$}) & $1.2$\\
\multicolumn{1}{l}{Termination reward components ($r^{ter}_{t}$):} & \\
Reward for reaching the finish (\texttt{dist\_episode\ >\ 3900}) & $+100$\\
Punishment for leaving the track (\texttt{|track\_pos|\ >\ 1.2}) & $-10$\\
Punishment for turning back (\texttt{angle\ <\ 0}) & $-10$\\
Punishment for damage the car (\texttt{damage\ >\ 0}) & $-10$\\
Punishment for progress backwards (\texttt{progress\ <\ 0}) & $-10$\\
Punishment for low progress (\texttt{timestep\ >\ 500\ AND\ episode\_reward < 0}) & $-10$\\[2pt]
\hline
\end{tabular}
\end{center}
\label{table1}
\end{table}

\begin{table}[!h]
\caption{Training hyperparameters for Stable Baselines PPO, and the TORCS simulator environment}
\begin{center}
\begin{tabular}{r@{\quad}rl}
\hline
\multicolumn{1}{l}{\rule{0pt}{12pt}
                   Parameter}&\multicolumn{2}{l}{Value}\\[2pt]
\hline\rule{0pt}{12pt}
Learning rate - with decay  &   $[1:2.5\cdot10^{-4}, 0: 0.5\cdot10^{-4}]$& \\
Maximum training steps & $1.5\cdot10^9$ \\
Discount factor  &     $0.995$& \\
Entropy coefficient  &   $0.01$& \\
Value function coefficient  &   $0.5$& \\
Policy clip range  &   $0.2$& \\
Value function clip range  &   $0.2$& \\
Environment instances & $24$& \\
Batch size & $512$& \\
Agent-Environment interaction timestep ($ts$) & $0.05$ sec\\
\multicolumn{1}{l}{TORCS parameters: } & \\
Used track name  &   brondehach& \\
Used car model name  &     155-DTM& \\
\texttt{ASR\_ON} & False& \\
\texttt{ASR\_ON} & False& \\
\multicolumn{1}{l}{Display mode constants:} & \\
\texttt{RM\_DISP\_MODE\_PYTORCS\_FIXFPS} & $8$& \\
\texttt{RM\_DISP\_MODE\_PYTORCS\_DISPLAY} & $16$& \\
\texttt{RM\_DISP\_MODE\_PYTORCS\_CAPTURE} & $32$& \\
Observation resolution & 84x84x4, grayscale &\\
Timestep in physics simulation & $0.002$ sec\\[2pt]
\hline
\end{tabular}
\end{center}
\label{table2}
\end{table}

\end{document}